\documentclass{article}
\usepackage{ijcai20}

\usepackage{times}
\usepackage{hyperref}
\usepackage{graphicx}
\usepackage{amssymb,amsmath}
\usepackage{amsthm}
\usepackage{booktabs}
\usepackage[linesnumbered,vlined,ruled,commentsnumbered]{algorithm2e}

\usepackage{subcaption}
\usepackage{tikz}
\usetikzlibrary{plotmarks}
\usepackage[switch,pagewise]{lineno}
\usepackage{color}

\usepackage{multirow}
\usepackage[normalem]{ulem}
\useunder{\uline}{\ul}{}

\title{Circuit Routing Using Monte Carlo Tree Search and  Deep Neural Networks}

\author{
\vskip -1em
Youbiao He and Forrest Sheng Bao
\affiliations
Dept. of Computer Science,
Iowa State University, Ames, IA 50014
\emails
\{yh54, fsb\}@iastate.edu
}

\date{}

\begin{document}

\maketitle

\begin{abstract}
Circuit routing is a fundamental problem in designing electronic systems such as integrated circuits (ICs) and printed circuit boards (PCBs) which form the hardware of electronics and computers. Like finding paths between pairs of locations, circuit routing generates traces of wires to connect contacts or leads of circuit components. 
It is challenging because finding paths between dense and massive electronic components involves a very large search space. Existing solutions are either manually designed with domain knowledge or tailored to specific design rules, hence, difficult to adapt to new problems or design needs. Therefore, a general routing approach is highly desired. In this paper, we model the circuit routing as a sequential decision-making problem, and solve it by Monte Carlo tree search (MCTS) with deep neural network (DNN) guided rollout. 
It 
could be easily extended to routing cases with more routing constraints and optimization goals. Experiments on randomly generated single-layer circuits show the potential to route complex circuits. The proposed approach can solve the problems that benchmark methods such as sequential A* method and Lee's algorithm cannot solve, and can also outperform the vanilla MCTS approach. 
\end{abstract}

\section{Introduction}\label{sec:intro}
Electrical and electronic devices in our lives, such as computers or phones, consist of many integrated circuits (ICs, also known as chips) and printed circuit boards (PCBs).
An IC or a PCB contains many physical wires that connect the components in or on it. 
A correct layout of such wires is important to the functionality and performance of the circuit. 
The problem of \emph{routing} in circuit design is hence to find the proper paths to place those wires to achieve expected connectivity with certain constraints, one of which is that wires carrying different signals should not intersect. Many Electronics Design Automation (EDA) software tools have been developed for this purpose. 
\begin{figure}[!htbp]
    \centering
    \includegraphics[width=0.7\columnwidth]{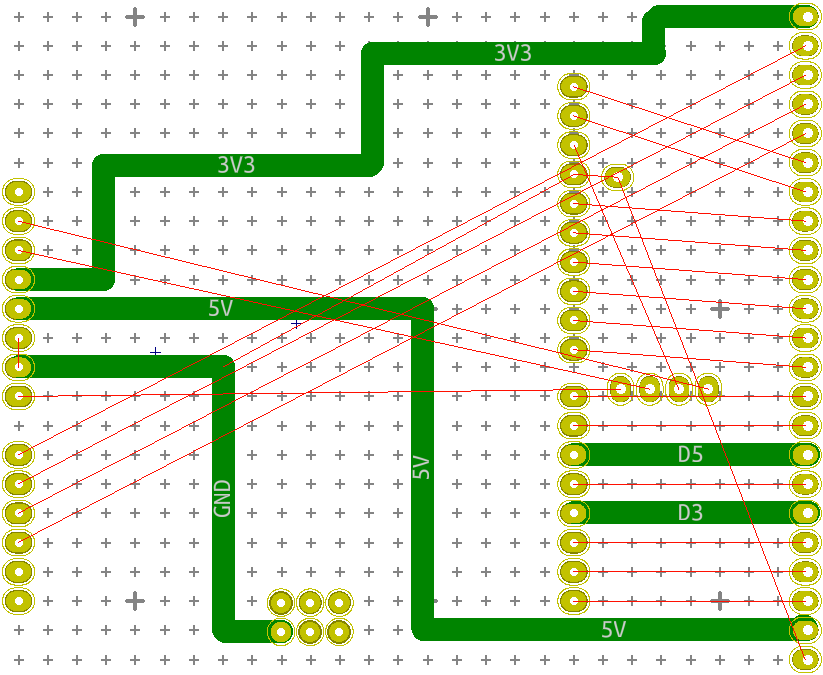}
    \caption{A PCB under routing in KiCAD. Red thin lines indicate nets to be connected and yellow pads are pins of nets. Green traces are finished paths connecting nets. Plus signs are grid intersections. }
    \label{fig:my_label}
\end{figure}

Circuit routing is computationally expensive and challenging. 
A PCB, such as the motherboard of a smartphone, can easily contain thousands of pins (a \emph{pin} is a contact, or metal lead, of an electronic component) and over ten layers, making manual design extremely time-consuming~\cite{5227133,Coombs_2007}. 
A 2012 chip could contain ``about 100 kilometers of copper wiring'' inside of it\footnote{\url{https://www.technologyreview.com/s/428466/making-wiring-that-doesnt-trip-up-computer-chips/}}.
On top of the problem size, circuit routing is NP-hard. 

Complex routing problems are usually solved in two stages: 
first global routing and then detailed routing~\cite{
handbook_routing,CHEN2009687,kahng2011vlsi}. 
Global routing routes on large circuit blocks, called G-cells, with relaxations on many constraints, while detailed routing generates wires of proper shapes and positions based on results from global routing. 
However, the two stages do not always couple well. 
For example, it is difficult for the wire density in global routing result to match the routability of the downstream detailed router~\cite{6241568}. 
As a result, a low-congestion global routing result may lead to subsequent detail routing difficult.
The miscoupling of the two stages becomes more profound as design rules evolve rapidly~\cite{Shi2017}. 
The distinction between the two stages also results in complex EDA software that is difficult to maintain and evolve coherently.

In addition, routing algorithms are dominantly manually crafted based on domain knowledge. 
Developing new algorithms is laborious.
Manual updates are always needed when new design constraints and goals arise. 
Many of these algorithms are based on heuristics which cannot guarantee solutions. For example, 
the order of nets typically affects the routing results~\cite{handbook_routing} and 
some routable problems may become insolvable in certain heuristics~\cite{zhang2016study}.

Therefore, an end-to-end and general routing approach that does not depend on domain knowledge is highly desired. 
In this paper, we view circuit routing as a sequential decision process that adds a vertex (or edge) onto a grid graph in each step, just like placing a stone in the game Go, which is suitable to be solved by Monte Carlo tree search (MCTS)~\cite{MCTS_survey}.
To better guide the search process, we use a deep neural network (DNN) that embeds the routing state.
Unlike previous approaches, our MCTS-based approach does not require domain knowledge nor specialized heuristics, and constraints and goals can be altered by revising the reward function without changing the algorithm itself.
Hence, it has lower implementation and maintenance costs, and the computer can improve its skills as it routes more and more circuits.

However, it is not trivial to apply the vanilla MCTS algorithm for circuit routing. 
On one hand, the average-reward based upper confidence bound for trees (UCT)~\cite{UCT} in vanilla MCTS does not correspond to the optimal paths in the routing problem.
On the other hand, the search space of routing is too large for random search. 
Hence, we propose a search algorithm that combines Monte Carlo simulation with policy networks. 
First, a maximum reward-based UCT~\cite{Monte_Mario,Max_UCT} is used to represent the best search result for an optimal path. 
Second, to reduce the search space, we combine the depth first search (DFS) and a trained neural network as the rollout policy of the MCTS algorithm. 


\section{Related Work}\label{sec:relatedwork}

In the design of integrated circuits (ICs) and PCBs, routing is a fundamental and challenging problem for decades. For the complex, large-scale circuit designs, circuit routing is usually solved by a two-stage approach, global routing followed by detailed routing~\cite{CHEN2009687}. 

There have been multiple approaches to global routing, such as force-directed routing~\cite{sequencerouting}, region-wise routing~\cite{HU20011}, and rip-up and reroute~\cite{ripup-reroute}.
\cite{7459314} proposes a procedure to improve the distribution of congestion without increasing the total overflow, and ~\cite{Shi2017} proposes a framework to analyze track congestion at the global routing stage. In addition, machine learning techniques have been used in subproblems in global routing, such as predicting routing congestion~\cite{6974668} or routability~\cite{7274019,8587655} with supervised learning.
A recent study models global routing as a deep reinforcement learning (DRL) problem~\cite{RL_routing},  where a deep Q-network (DQN) is implemented. Unlike our DNN-based MCTS, where a decision is made based on the results of many simulations using a trained policy, the DQN makes the decision according to only the output of its policy. However, a trained policy by DQN cannot guarantee to find a satisfying solution for an unseen and complex circuit.

Given global routing results, detailed routing generates wires of proper shapes and positions. One of the most popular algorithms is channel routing~\cite{deutsch1988dogleg,68407,channel_new}. It normally decomposes the routing region into routing channels and connects wires within these channels~\cite{CHEN2009687}. 
However, global routing and detailed routing do not couple well for complex routing problems~\cite{Shi2017}, and
the implementation of this two-stage approach usually results in complex software systems difficult to maintain and verbose workflow that slows down the circuit development.

\begin{figure}[!htbp]
\centering
\begin{subfigure}{0.17\textwidth}
  \centering
  \includegraphics[width=\textwidth]{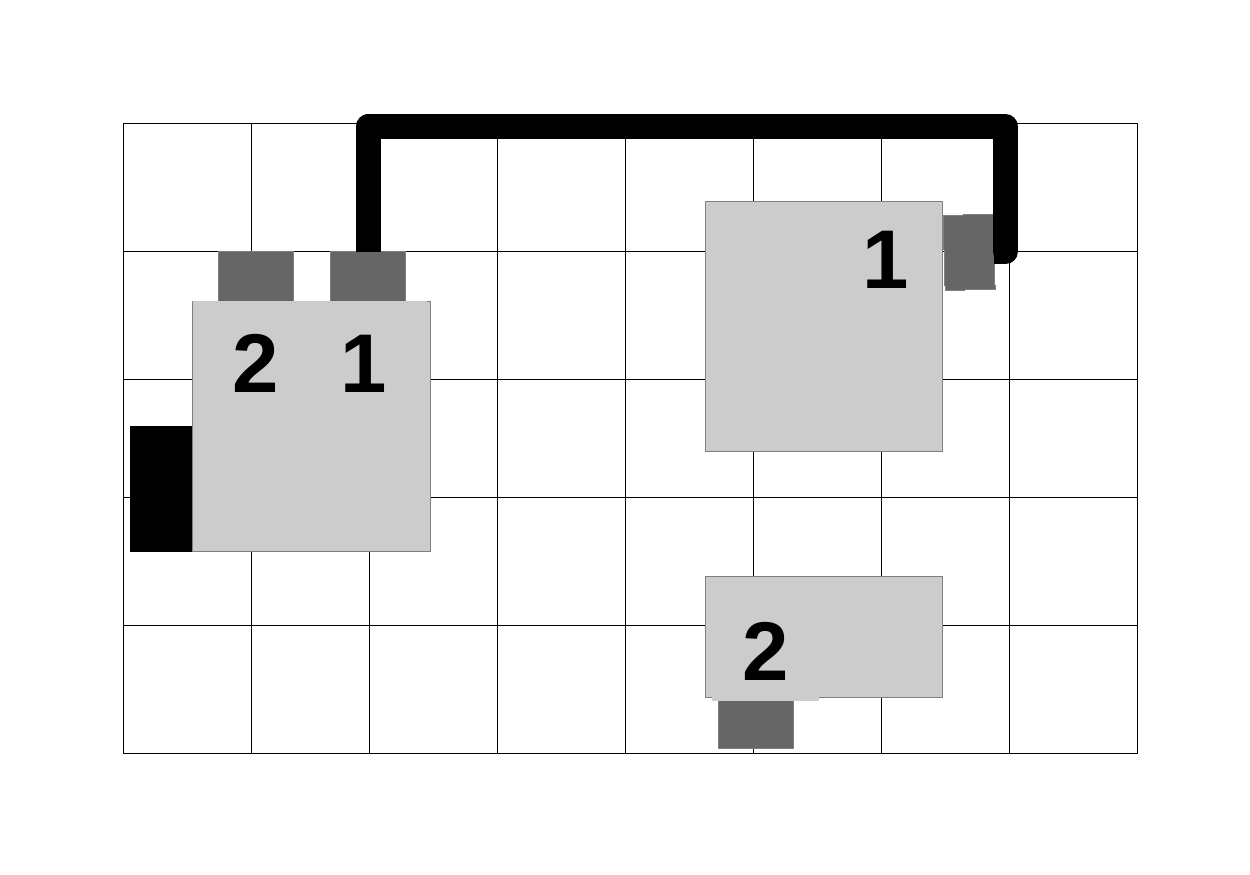}
  \label{fig:netorder_a}
\end{subfigure}%
\begin{subfigure}{0.17\textwidth}
  \centering
  \includegraphics[width=\textwidth]{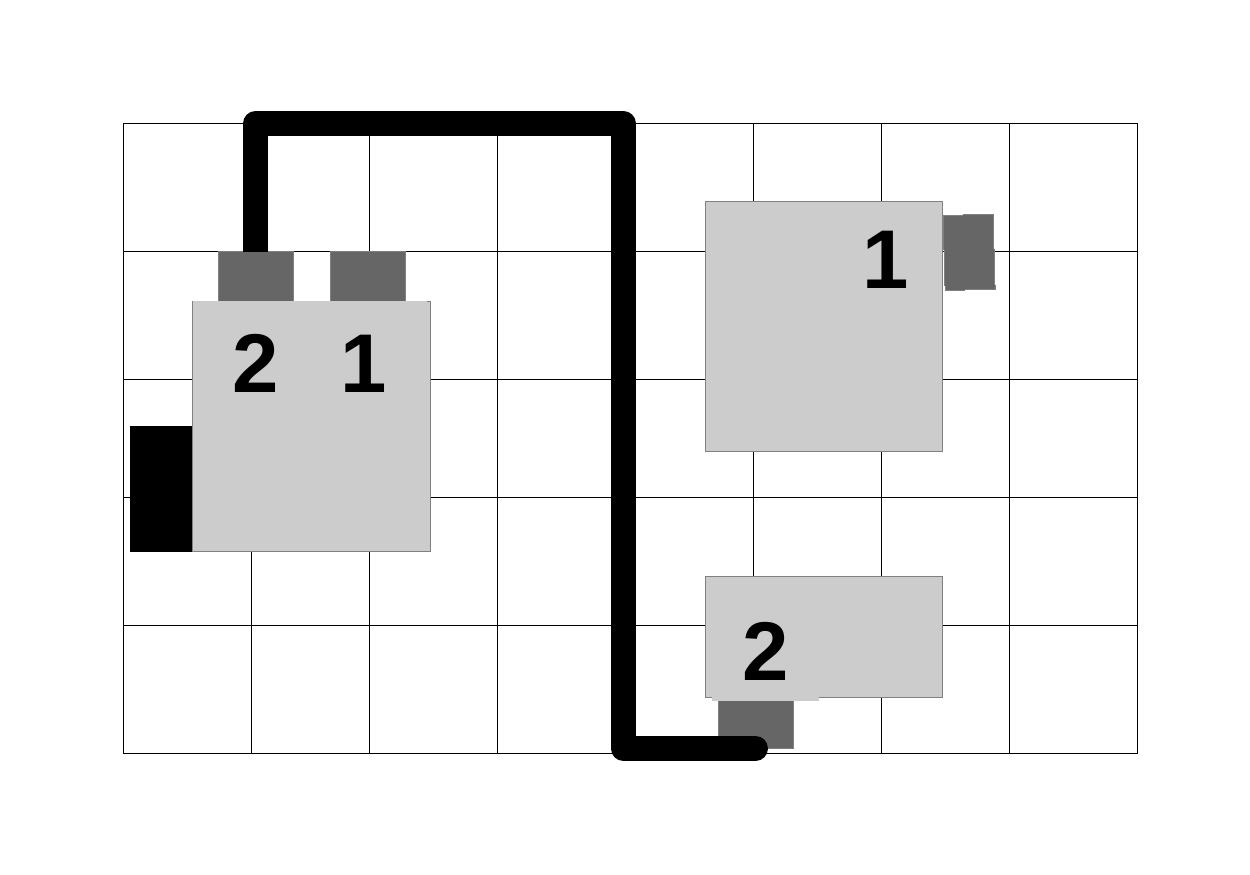}
  \label{fig:netorder_b}
\end{subfigure}%
\begin{subfigure}{0.17\textwidth}
  \centering
  \includegraphics[width=\textwidth]{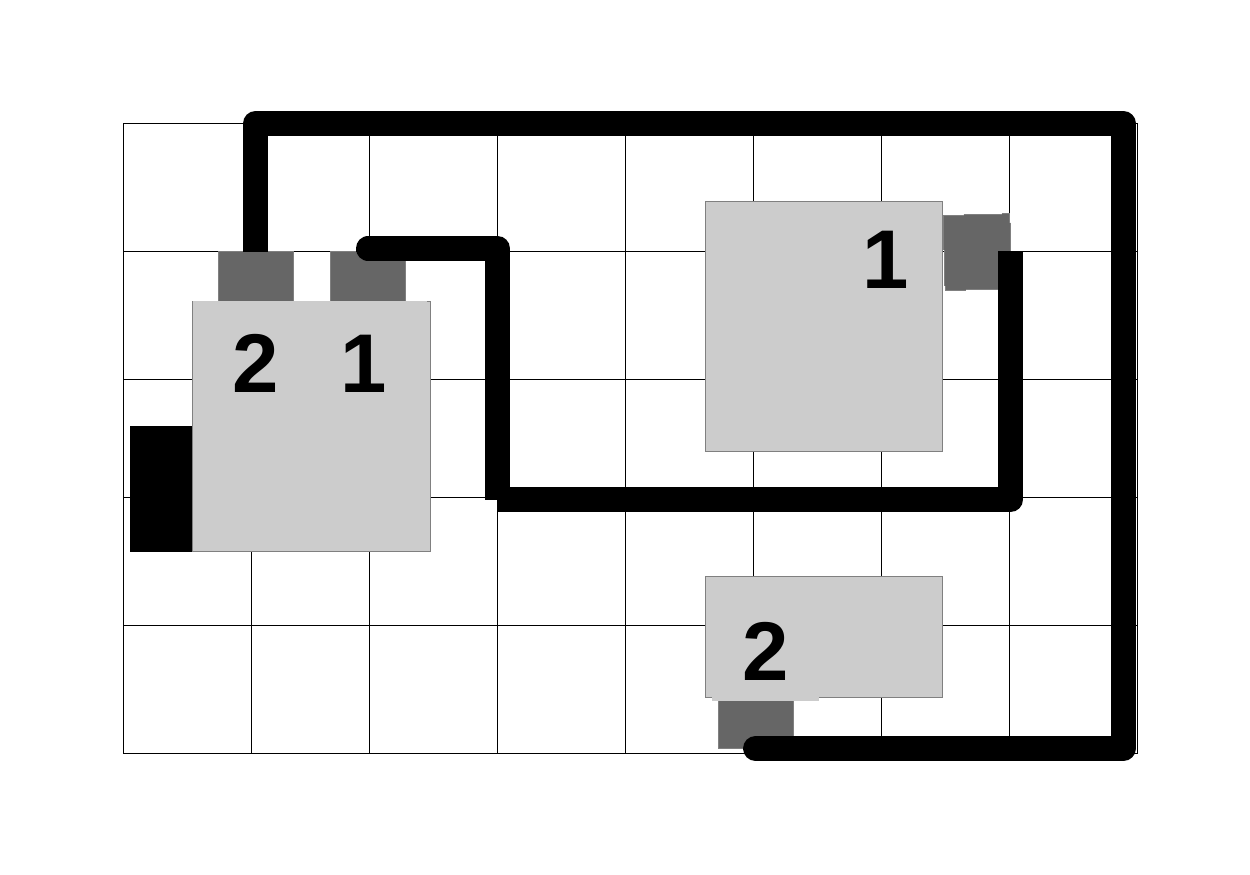}
  \label{fig:netorder_c}
\end{subfigure}
\vskip -2em
\caption{Connecting nets in greedy shortest paths may leave a routing problem insolvable. The dark black block means a non-routable region, i.e., no paths can  go through it.}
\label{fig:netorder}
\end{figure}

For designs with only one or two metal layers and a small number of nets, routes are usually constructed directly rather than in the two-stage approach~\cite{kahng2011vlsi}. This category of routing is called area routing. The most popular area routing approaches are  sequential routing based on the maze routing algorithms such as A* algorithm and Lee’s algorithm~\cite{AstarandLee,238611,124813,724497}. 
However, they are very sensitive to the net orders~\cite{zhang2016study}, and in some cases, the greedy search techniques used by them may make a solvable problem insolvable.
Figure~\ref{fig:netorder} shows solving a 2-net routing problem in 3 approaches. Connecting  net 1 first in shortest path leaves no paths for wiring net 2 (Figure~\ref{fig:netorder}, Left) and vice versa (Figure~\ref{fig:netorder}, Center). But a better routing solution can be achieved if the earlier path does not block the path of the later net (Figure~\ref{fig:netorder}, Right).
Although our approach is still sequential routing, 
as to be explained theoretically and empirically later in this paper, our approach is not prone to net ordering.

Path finding has also being intensively studied in robotics and video games~\cite{Sturtevant_2014,sven2020}. However circuit routing differs from them in that paths are allowed to intersect in them. This makes circuit routing much more difficult than many multi-agent path finding problems. 

\section{Problem Formulation}\label{sec:circiutRouting}
Similar to navigation (e.g., using Google Maps on your smartphone in your car), circuit routing is a path search problem. 
One can imagine circuit routing as given many pairs of locations, finding non-intersecting paths that connect these pairs on a map. 
For simplicity, our discussion is limited to 2D spaces (thus, single-layer circuits), but the solution can be easily expanded to 3D spaces for multi-layer circuit routing. 

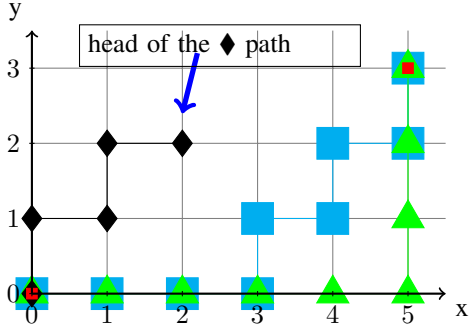
\begin{figure}[!htbp]
    \centering
\begin{tikzpicture}

\draw [thin, gray] (0,0) grid (5.5,3.5);

\draw[->,blue, line width = 2](2.2,3.2) -- (2,2.4) ;
\node[draw, text width=3.5cm] at (2.5,3.3) {head of the  $\blacklozenge$ path};

\draw [cyan] plot [mark size = 6, mark=square*] coordinates {(0,0) (1,0) (2,0) (3,0) (3,1) (4,1) (4,2) (5,2) (5,3) };
\draw [green] plot [mark size = 7, mark=triangle*] coordinates {(0,0) (1,0) (2,0) (3,0) (4,0) (5,0) (5,1) (5,2) (5,3) };
\draw [black] plot [mark size = 5, mark=diamond*] coordinates {(0,0) (0,1) (1,1) (1,2) (2,2) };

\node[mark size=2,color=red] at (0,0) {\pgfuseplotmark{square*}};
\node[mark size=2,color=red] at (5,3) {\pgfuseplotmark{square*}};

\draw[thick,->] (0,0) -- (5.5,0) node[anchor=north west] {x};
\draw[thick,->] (0,0) -- (0,3.5) node[anchor=south east] {y};
\foreach \x in {0,1,2,3,4,5}
  \draw (\x cm,1pt) -- (\x cm,-1pt) node[anchor=north] {$\x$};
\foreach \y in {0,1,2,3}
    \draw (1pt,\y cm) -- (-1pt,\y cm) node[anchor=east] {$\y$};

\end{tikzpicture}
    \caption{Three paths, two (
    \textcolor{cyan}{\rule{0.7em}{.7em}--\rule{0.7em}{.7em}}
    and 
    \textcolor{green}{$\blacktriangle-\blacktriangle$}
    ) of which connect a net (two pins in small red squares \textcolor{red}{\rule{0.5em}{.5em}}) while one ($\blacklozenge-\blacklozenge$) of which does not. }
    \label{fig:leaf_bead}
\end{figure}


Circuit routing starts from constructing a graph $G=(V, E)$
via sampling on a uniform Cartesian grid. 
For every intersection on the grid, a vertex is created into $V$.
For every vertex $v_1\in V$ and each of its 1-hop neighbors $v_2\in V$, an edge $e = \{v_1, v_2\}$ is added into $E$.

An electrical circuit may contain many non-routable zones such as being part of a drill hole, in the clearance zone of a part, or out of the boundary of the IC or PCB.
The set $O$ consists of all vertices falling into non-routable zones.
In addition, an electrical circuit may also contain many sets of electrically equivalent pins, e.g., all ground pins form a set. All vertices in the same set of electrically equivalent pins form a \emph{net} $N=\{v_{n1}, v_{n2},\dots\}\subseteq V$.
In this paper, each net contains two and only two vertices because a multi-pin net can be decomposed into multiple two-pin nets easily using a minimum spanning tree (MST) or a rectilinear Steiner tree (RST)~\cite{rst,HU20011}, and to simplify the problem, nets do not share any vertex.

A two-pin net $N=\{v_{s}, v_{e}\}$ is \emph{connected} by a path\footnote{In graph theory, a path could be a sequence of vertices, a sequence of edges, or a sequence of alternating vertices and edges. In this paper, we adopt the first one for simplicity. Given a path $P=[v_1, v_2, \dots, v_n]$ defined as a sequence of vertices, there is an edge $\langle v_i, v_{i+1} \rangle \in E$ for all $i\in [1..n-1]$.} 
$P=[v_1, v_2, \dots, v_n]$
if $v_s=v_1\text{ and } v_e=v_n$, and $P \cap O = \emptyset$.
A two-pin net is partially connected if only one of the pins is on the path, e.g., $\blacklozenge$'s in Figure~\ref{fig:leaf_bead}. 
Given a path $P$ that partially connects a net, its last vertex is called a \emph{head}, e.g., the $\blacklozenge$ at $(2,2)$ in Figure~\ref{fig:leaf_bead}. 
Note that a net can be connected by multiple paths (e.g., $\blacksquare$'s and $\blacktriangle$'s both connect the net of \textcolor{red}{\rule{0.5em}{.5em}} in Figure~\ref{fig:leaf_bead}). 
A set $\mathcal{P}$ of paths are \emph{non-intersecting} if  they do not share any vertex. 
For example, the paths for the power and ground signals cannot intersect, otherwise the shortcut will fry the circuit. 

A \emph{routing problem} is, given a graph $G=(V, E)$, a set of obstacles $O\subseteq V$, and a set $\mathcal{N}=\{N_1, N_2, \dots, N_k\}$ of $k$ nets, 
to find a non-intersecting set $\mathcal{P}$ of $k$ paths such that every net is connected by one and only one path. 
A routing problem is \emph{unsolvable}, if such a set $\mathcal{P}$ of paths does not exist.


Quite often, in circuit routing, we have constraints and goals, such that certain solutions are not allowed and certain solutions are better than others. In experiments of this paper, we just consider a simple but common goal that the total length of paths should be minimized, i.e., $\min \sum\limits_{p\in\mathcal{P}} |p| = \left | \bigcup \mathcal{P} \right | $. 







\section{Methods}\label{sec:methodology}

\subsection{MCTS-based circuit routing}

Given a net, our approach constructs a path from one of the two pins of the net, expanding it by one vertex each step. 
A special variable $h$, representing the head of the path under expansion, is tracked.
At each step, neighbors (excluding those already visited, obstacles or pins in other nets) of the head $h$ are evaluated using MCTS and one of them is picked as the new $h$ from which the path expansion continues. 
Similarly in the game Go, 
the AI player calls MCTS at each of its turns to determine the best grid intersection to add a stone. 
MCTS is a search algorithm for deciding optimal actions to take at discrete steps. It has been widely used in the domain of game playing, such as Go~\cite{Silver_2016}, chess and shogi~\cite{silver2017mastering}, and Atari Games~\cite{guo2016deep}.

\begin{figure}[!htbp]
    \hskip -1em
    \includegraphics[width=1.15\columnwidth]{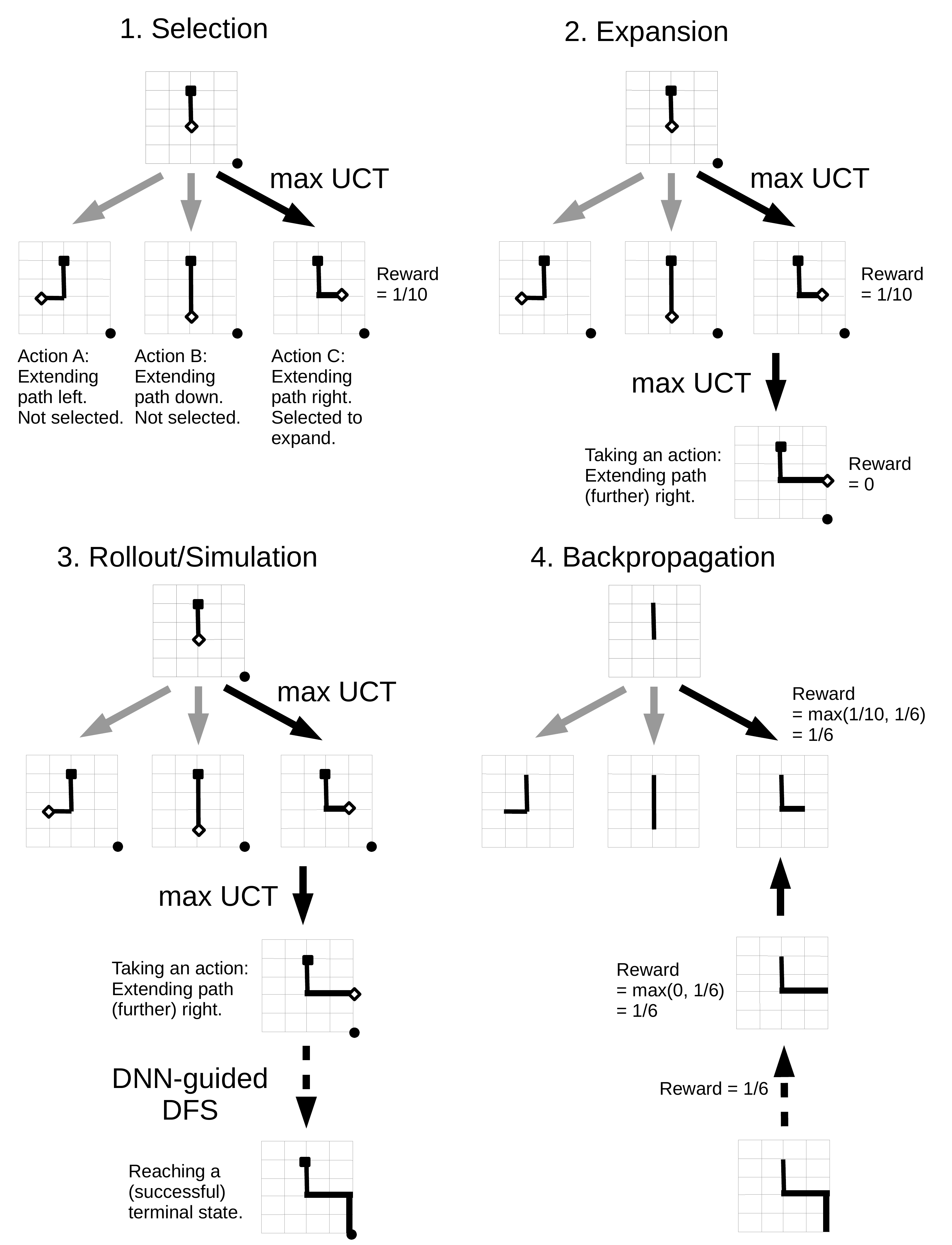}
    \caption{Four stages of an MCTS iteration when routing an example circuit of only one net, from the square to the dot. The head at any state is marked in a void diamond. Nodes are selected in max UCT. Rewards are calculated in Eqn.~(\ref{eqn:reward}) and backpropagated in max UCT. 
    }
    \label{fig:MCTS}
\end{figure}

Each MCTS call builds a search tree iteratively until a predefined number of iterations is reached, at which point the search is halted and the child node of the root that maximizes the reward is returned and corresponds to the best action to take.
As illustrated in Figure~\ref{fig:MCTS}, each MCTS \emph{iteration} consists of four stages: selection, expansion, rollout/simulation, and backpropagation~\cite{MCTS_survey}.
Each tree node represents a state and 
has children corresponding to expanding the path toward different directions.
The reward for each state is updated through backpropagation after the rollout in each iteration.

In our approach, a \emph{state} is defined as a matrix $M$ whose element at the $i$-th row and the $j$-th column is:
\begin{equation}
\label{eqn:state}
  M_{ij}=\begin{cases}
    i_n, & \text{if } v_{ij}==h,  \\
    -1, & \text{else if } v_{ij}\in O\cup \bigcup \mathcal{P}, 
    \\
    n, & \text{else if } v_{ij}\in N_n \text{ and } v_{ij}\not \in \bigcup \mathcal{P},\\

    0, & \text{otherwise.} \\
  \end{cases}
\end{equation}
where $v_{ij}$ is the vertex at point $(i,j)$ on the grid, $O$ is the set of obstacles, 
$n\in [1..k]$ is the index of nets, 
$i_n$ is the index of the net being connected, 
$h$ is the head of the path that partially connects this net,  
and $\mathcal{P}$ is the set of all paths including the one that partially connects a net. Note that our routing approach connects nets sequentially. Hence, there is only one head at any time. The intuition behind our state definition is that all vertices that should not be considered have the value $-1$, and all vertices that are available to be part of a path have the value $0$.

\begin{figure}[!htbp]
    \centering
    \includegraphics[width=0.1\textwidth]{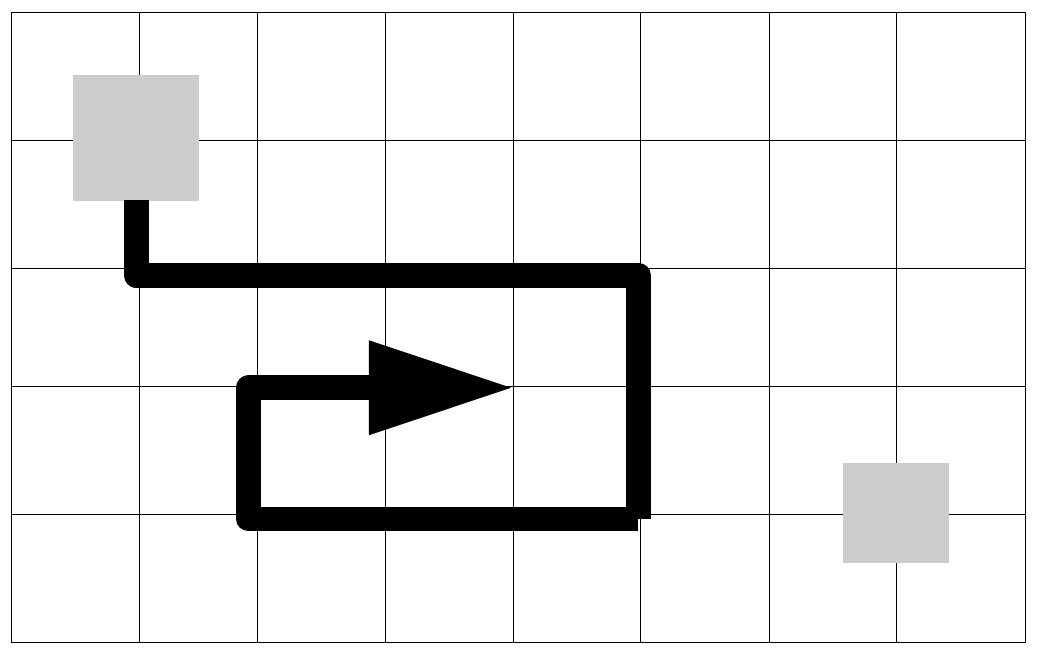}
    \caption{An example dead end where all actions are illegal.}
    \label{fig:dead_end}
\end{figure}

At each state, there are four \emph{actions} to consider, denoted as ``up,'' ``down,'' ``left,'' and ``right,'' each of which corresponds to expanding the path to one of the four 1-hop neighbors of the head $h$. 
An action is \emph{illegal} if the vertex corresponding to it belongs to $O$, $\bigcup {P}$, or $N_m$ such that $m\in [1..k]$ but $n\not = m$. Note that among all the four possible actions, illegal actions are redundant and will not be visited at all. 
The \emph{reward} of a \emph{terminal state}, corresponds to a failed or succeed routing, is calculated as 
: 
\begin{equation}
\label{eqn:reward}
  R=\begin{cases}
    \frac{1}{|V|}, & \text{if routing fails}. \\
    \frac{1}{|\bigcup \mathcal{P}|}, & \text{if routing succeeds}.
  \end{cases}
\end{equation}
where $|\bigcup \mathcal{P}|$ is the total length of paths (the number of vertices along the paths), and $|V|$ is the total number of vertices in the graph $G$. The routing \emph{succeeds} if all nets are connected. 
The routing \emph{fails} if there are no legal actions to take (such as the dead end shown in Figure~\ref{fig:dead_end}) and the net is not connected by the path under construction. 
Because $\bigcup \mathcal{P}\subseteq V$, the condition $\left |\bigcup \mathcal{P}\right |\leq |V|$ always holds. 
Note that this reward involves the optimization goal of our routing problem, which minimizes the total wire length. If the routing succeeds, a larger reward is returned by a shorter wire length. Hence, the algorithm is forced to find a path as short as possible.

Rewards propagate back to the root depending on the UCT function.
If using average UCT of vanilla MCTS, then the reward for a non-rollout (i.e., already on the tree) node is updated 
to the new average reward by counting the new reward in. 
If using maximum reward-based UCT~\cite{Monte_Mario,Max_UCT} (See Section~\ref{sec:actionEva}), the reward for a non-rollout node is updated to the new maximum considering the new reward. 


The overall picture of our approach is given in Algorithm~\ref{alg:algorithm_seq}. 
It routes one net each time. 
Although sequential, it will not be impacted by the net order because blocking the way of the later nets will cause a routing fail, thus a low reward. 
In experiments later,  the results also show that our approach can solve routing problems that traditional sequential routing approaches, such as sequential A* or Lee's algorithm \cite{AstarandLee,238611,124813,724497},  cannot due to net ordering. 

\begin{algorithm}[!htbp]
\SetAlgoLined
\For{$n \in [1..k]$}{
$h \sim \mathcal{U}(N_n)$

$P_n \leftarrow \emptyset$

\While{true}{
$P_n\leftarrow P_n\cup \{h\}$.

$v_{best} \leftarrow \texttt{MCTS-DNN}(M)$. 

\uIf{$v_{best}$ \text{is not None}}{

\uIf{$v_{best}\in N_n\setminus P_n$}{
$P_n\leftarrow P_n\cup \{v_{best}\}$.


break // $N_n$ is connected.
}
\Else{

$h \leftarrow v_{best}$



}

}
\Else{Routing fails. Return. }
}
}
\caption{MCTS-based circuit routing. }
\label{alg:algorithm_seq}
\end{algorithm}




MCTS uses two policies to guide the tree search. One is action evaluation policy, which is used in the stages of node selection and expansion. The other is rollout policy, which is applied in the rollout stage. 
However, these two policies in the vanilla MCTS cannot efficiently search in the huge space of circuit routing. 
In response, the maximum reward-based UCT method~\cite{Monte_Mario,Max_UCT} is implemented for action evaluation (Section~\ref{sec:actionEva}), and a DNN-based depth first search (DNN-DFS) method is proposed as the rollout policy (Section~\ref{sec:rollout_policy}).

\subsection{Action Evaluation}\label{sec:actionEva}
In the vanilla MCTS algorithm, 
a child node is selected in the selection and expansion stages if it maximizes the upper confidence bounds for trees (UCT) among its siblings:
\begin{equation}
\label{eqn:ucb}
    UCT = \overline{R}+C_p\sqrt{\frac{2\ln{n_p}}{n_c}}
\end{equation}
where $\overline{R}$ is the average reward from all the rollouts of a node $c$, $n_p$ is the number of times that $c$'s parent node has been visited, $n_c$ is the number of times the node $c$ has been visited, and $C_p>0$ (0.5 in our experiments) is an exploration parameter.
Unvisited children, with $n_c=0$ and thus $UCT=\infty$, have the largest possibility to be visited in the current iteration. 
Hence every child is guaranteed to be visited at least once.

However, it may be problematic to employ UCT for circuit routing. 
Given two actions, from their average rewards of multiple rollouts/simulations, we cannot tell which one has ever reached the greater maximal reward and thus the shorter shortest path, e.g., although  $\max(1,2,3)=3>\max(2,2,2)=2$,  $\overline{(1,2,3)}=\overline{(2,2,2)}=2$. 
Therefore, the maximum reward-based UCT scheme~\cite{Monte_Mario,Max_UCT} is implemented in this paper, which is defined as follows: 
\begin{equation}
\label{eqn:uct}
    R_{\max}+C_p\sqrt{\frac{2\ln{n_p}}{n_c}}
\end{equation}
where $R_{\max}$ is the maximum reward from rollouts. 
For the rest of this paper, the UCT action evaluation based on the maximum and average reward are denoted as Max-UCT and Avg-UCT, respectively. 



\subsection{Rollout policy}\label{sec:rollout_policy}
In the vanilla MCTS algorithm, the agent randomly selects actions 
in a rollout. 
But in the circuit routing, due to the large search space, 
a random rollout policy is very inefficient 
as only an extremely small amount of rollouts end up with successful routing results. 
To improve the search efficiency, we propose a DNN-guided depth-first search (DNN-DFS) as the rollout policy. 
The DNN improves the success rate by prioritizing nodes while the DFS saves time by avoiding cold-starting every rollout. 

\begin{figure}[!htbp]
    \centering
    \includegraphics[width=.9\columnwidth]{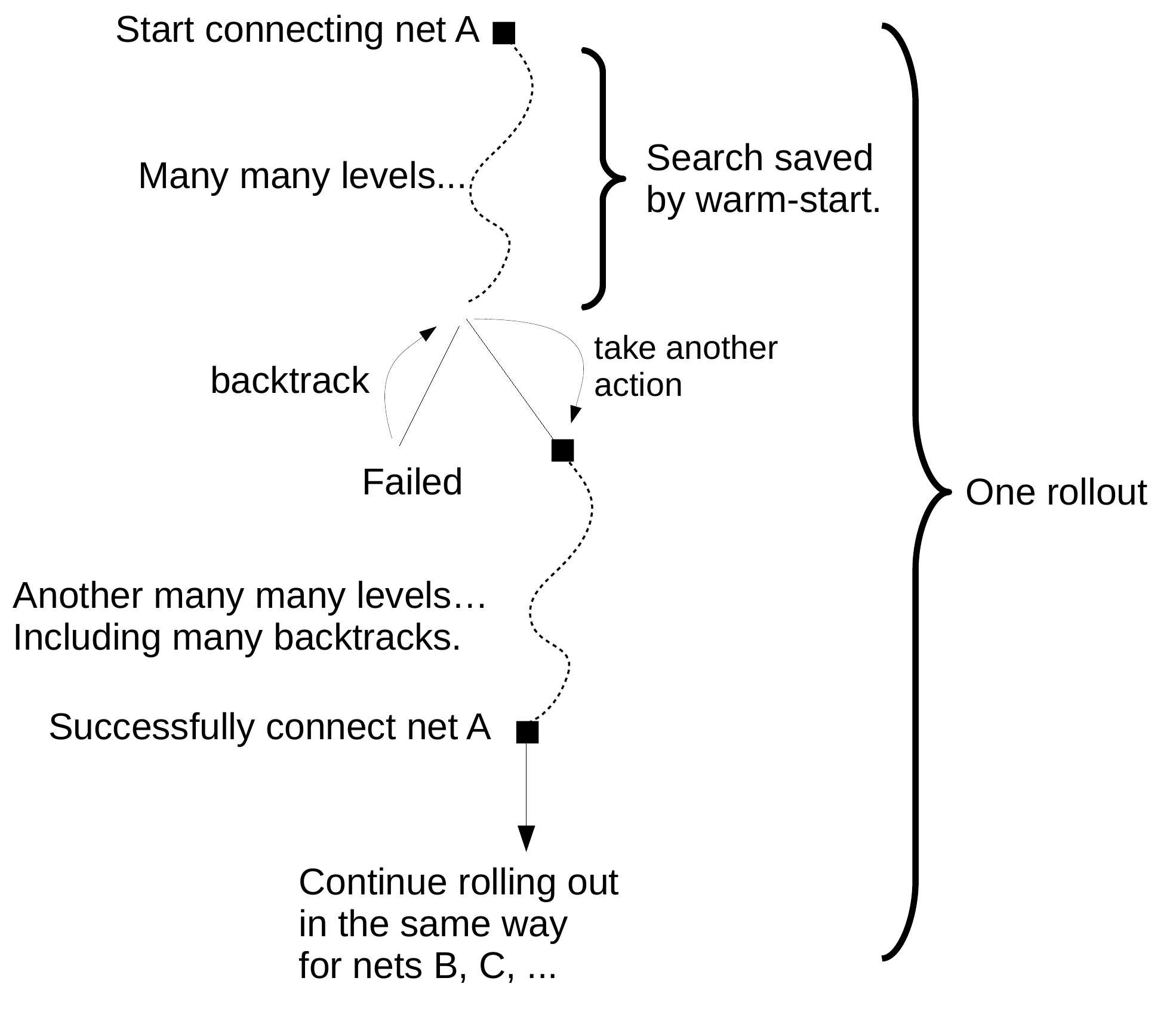}
    \caption{DFS-based rollout. }
    \label{fig:DNN_DFS_2}
\end{figure}

In the DFS scheme (Figure~\ref{fig:DNN_DFS_2}), if a rollout ends up at a failed routing, it does not terminate, but backtracks to continue the searching. 
The target to be searched is sequentially updated to be the nets to be connected. 
Once a net is connected (i.e., both pins of the net are visited),
the target is updated to the end vertex/pin of the next net. 
Note that a rollout can still end up with a dead end if connecting net is not connectable, for example, all the paths connecting this net are blocked by the routing of early nets. 

Embedding the states, the DNN predicts the probability of a successful routing by taking each of the candidate actions, and prioritize nodes in the search process based on the probabilities. 
The DNN is implemented by a convolutional neural network (CNN) consisting of one convolutional layer (32 5$\times$5 filters, ReLU activation), one max-pooling layer (2$\times$2 2D max-pooling), and two fully-connected layers (128 neurons with ReLU activation and 4 neurons with softmax activation). 
The output neuron corresponding to the best action should have the maximal activation. 
In the configuration of this paper, the DNN maps a 30-by-30 state matrix into a 1-by-4 vector, each element of which corresponds to expanding the path toward one direction from the head. 

To train this DNN, we create a dataset using 2000 randomly generated circuits (grid size 30-by-30) routed in the maze routing algorithm and manual routing. 
Taking a random vertex on a random path/net of a circuit, we can create a head using the vertex, a state such that all nets prior to the net are connected and the net is connected up to the head, and one correct action that extends the path toward the next vertex on the path. 
In the expect 1-by-4 output vector, the element corresponding to correct action is 1 and the rest 3 are all 0. 
However, such random sampling can lead to overfitting easily because two consecutive states are quite similar, differing by only one vertex. 
To increase the variety of samples, we take only one sample for each net, resulting in 9,459 samples. 
The training and test sets follow 80-20 split. 
The accuracies for training and testing are $86.60\%$ and $85.04\%$, respectively.

The output distribution
of the DNN cannot be directly used to guide rollout because some actions could be illegal. 
Hence the action probabilities are normalized using the sum of probabilities of all legal actions. 

\section{Experimental Results}\label{sec:results}

Our proposed method is examined using 30 randomly generated 1-layer circuits on a grid of $30\times 30$.
The routability of each circuit is guaranteed. The optimization goal is minimizing the total length of paths.  

\subsection{Qualitative analysis}

\begin{figure}[!htbp]
\centering
\begin{subfigure}{0.25\textwidth}
  \centering
  \includegraphics[width=0.95\textwidth]{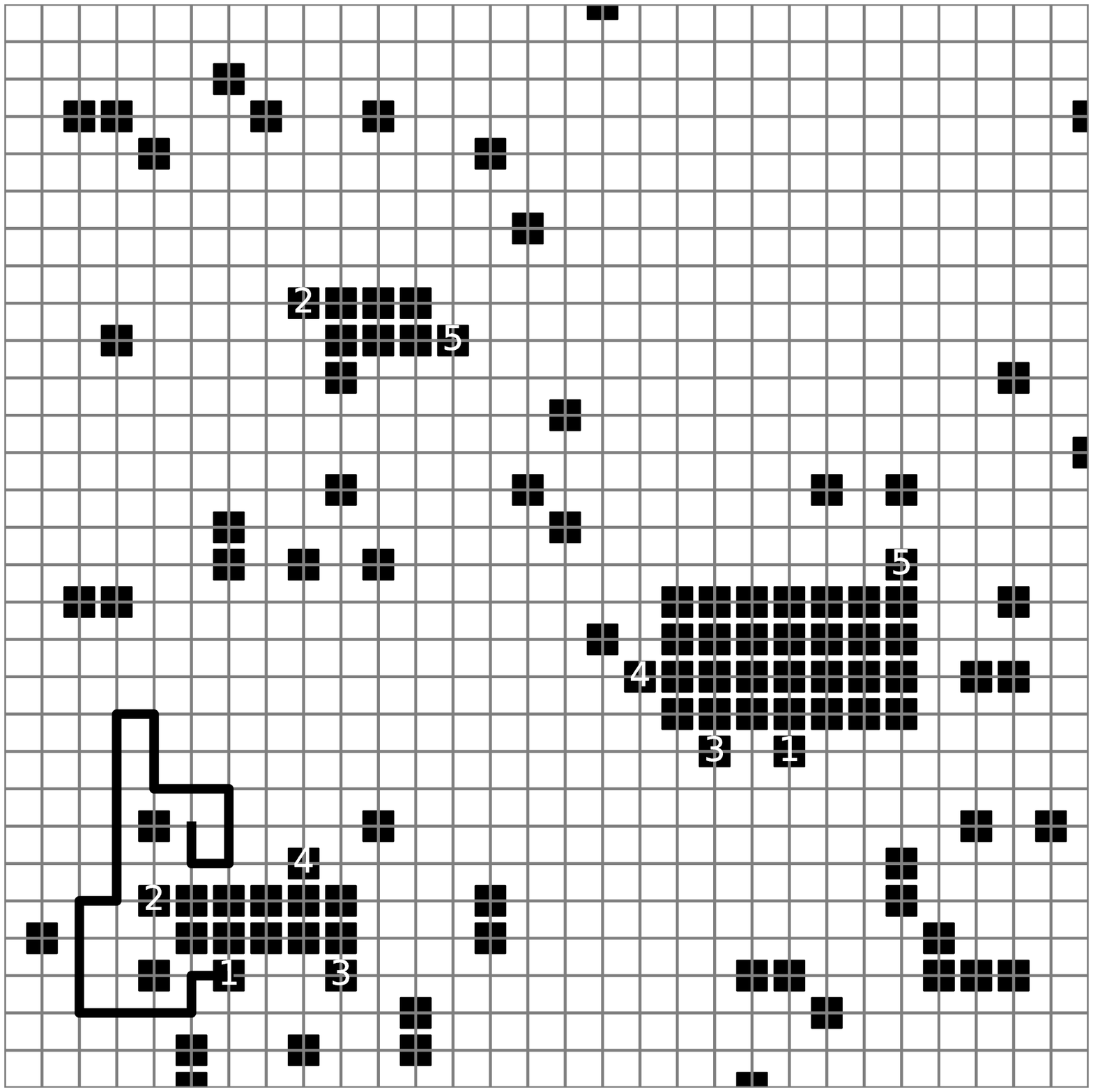}
  \caption{100 iterations.}
  \label{fig:route_100
  }
\end{subfigure}%
\begin{subfigure}{0.25\textwidth}
  \centering
  \includegraphics[width=0.95\textwidth]{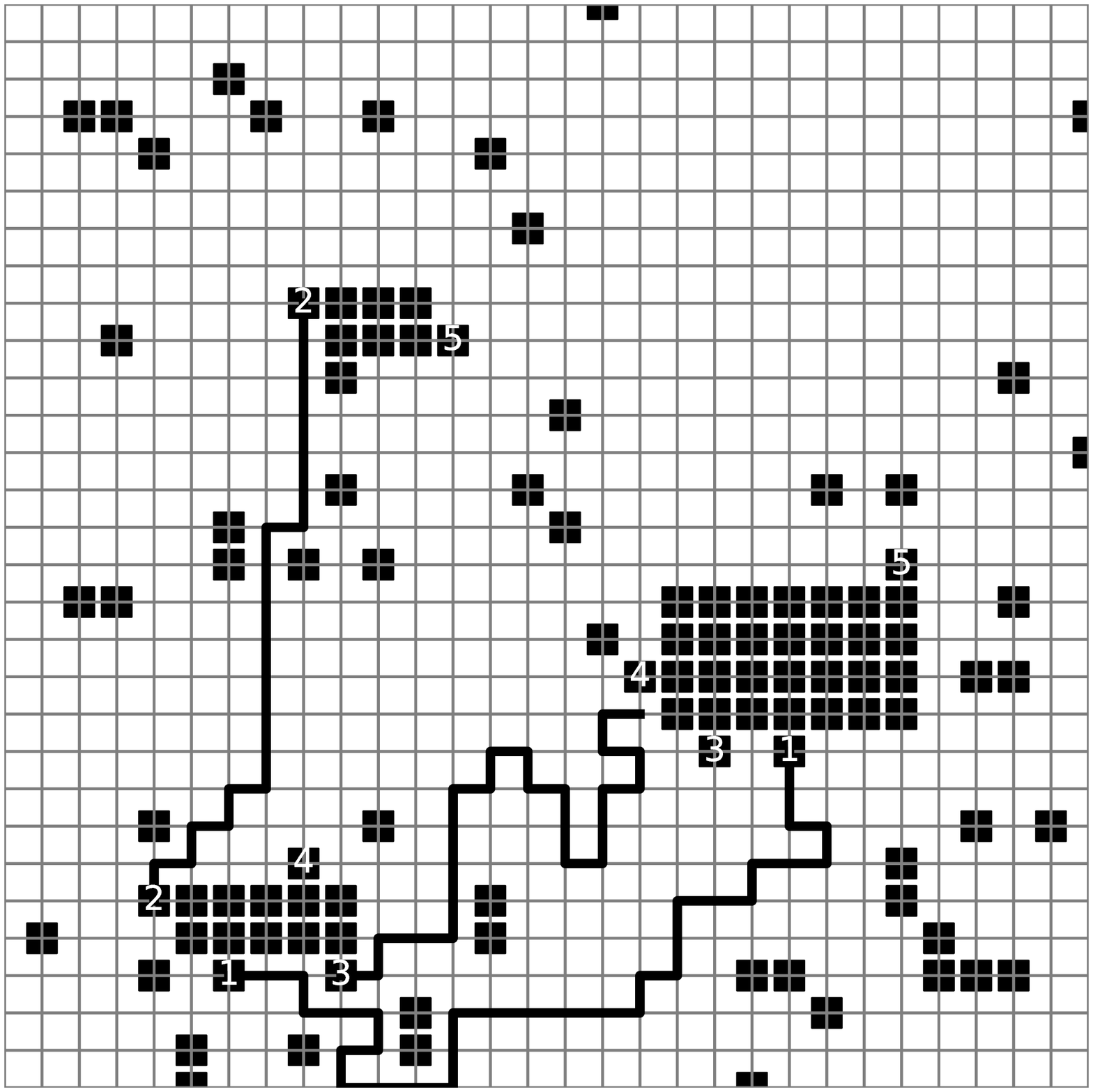}
  \caption{500 iterations.}
  \label{fig:route_500}
\end{subfigure}%
\\
\begin{subfigure}{0.25\textwidth}
  \centering
  \includegraphics[width=0.95\textwidth]{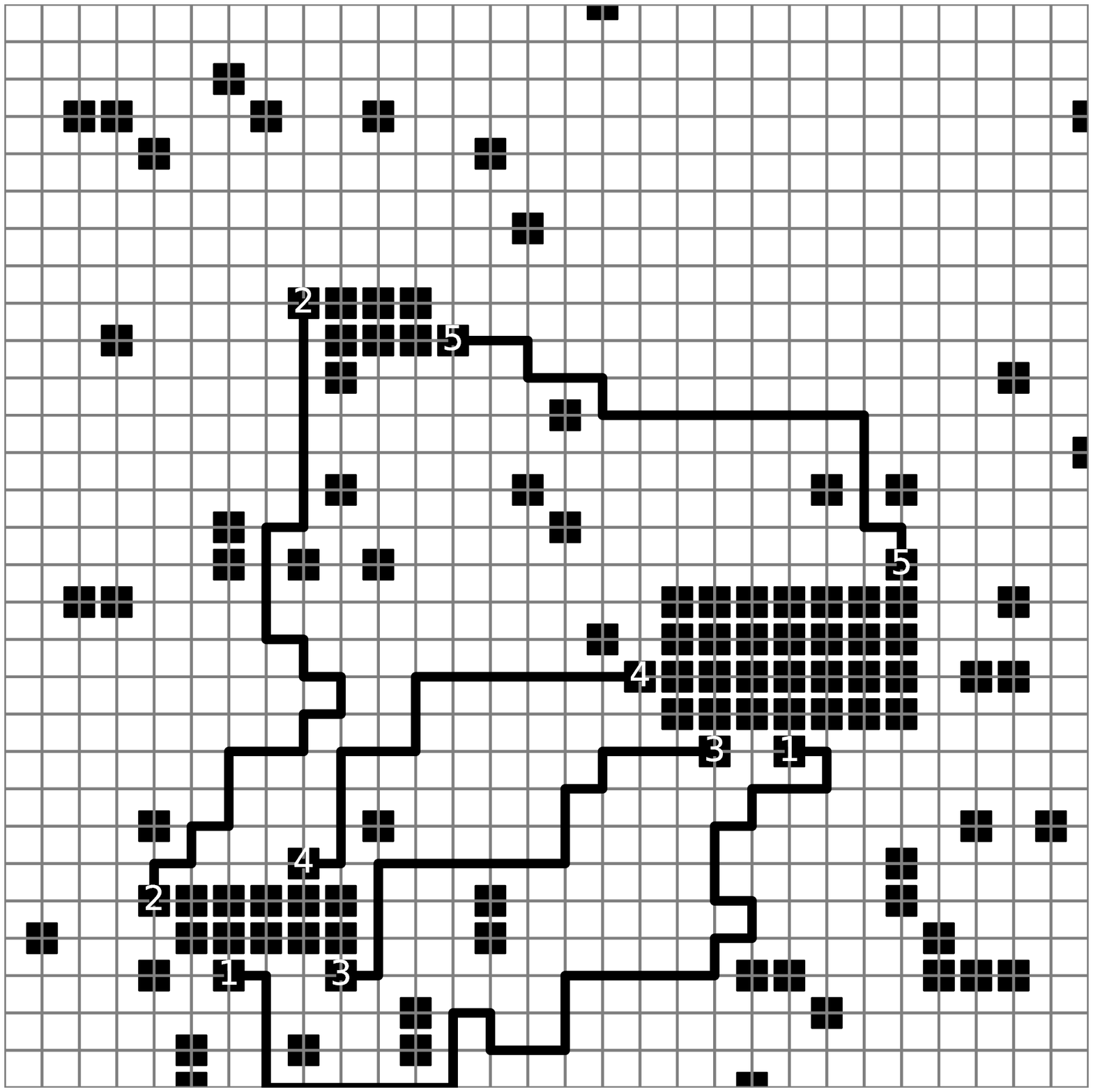}
  \caption{2000 iterations.}
  \label{fig:route_2000}
\end{subfigure}%
\begin{subfigure}{0.25\textwidth}
  \centering
  \includegraphics[width=0.95\textwidth]{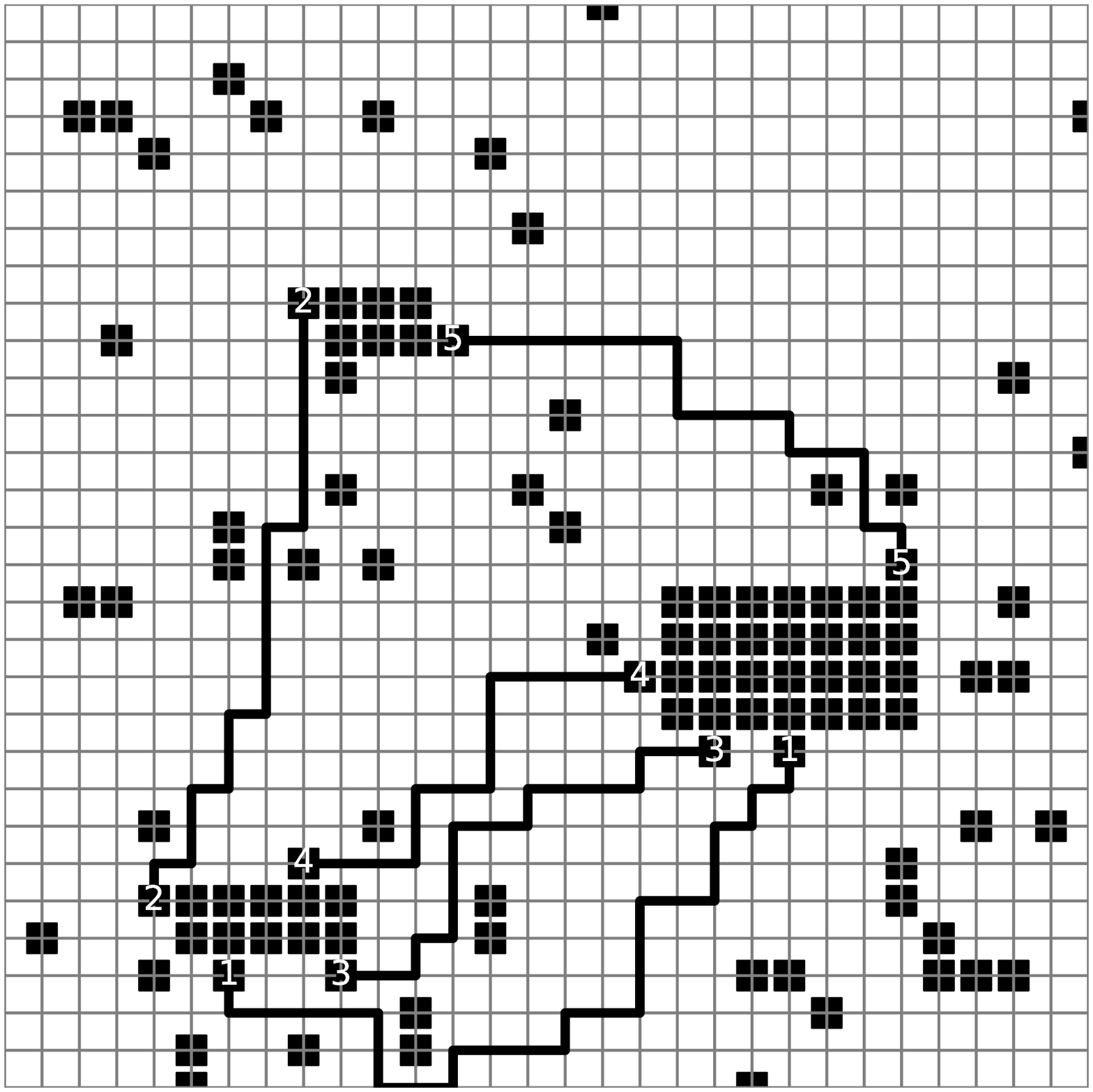}
  \caption{5000 iterations.}
  \label{fig:route_5000}
\end{subfigure}
\caption{Routing result gets better as the number of iterations increases. Obstacles are in solid black squares, nets are in numbered black squares, routed paths are in solid black lines.}
\label{fig:Routedboard_iters}
\end{figure}

Figure~\ref{fig:Routedboard_iters} shows an example of the routing with different numbers of iterations, in which the agent learns to improve the routing strategy as the iteration increases. With 100 iterations, none of the nets is successfully connected and the routing goes to a dead end. At 500 iterations, two of the nets are connected and the routing for net 3 goes to a dead end.
When the number of iterations reaches 2000, the agent learns the strategy to connect all nets, and with 5000 iterations, the shortest paths are found.

In many cases, the shortest paths cannot be guaranteed because the MCTS agent could make suboptimal actions, e.g., circling around.
This shows a major difference between routing and game playing where MCTS is also widely used. 
For example, in the game Go, a suboptimal move can be compensated by later moves.
But in circuit routing, 
a suboptimal action creates extra segments of wires that will forever stay on a path. 

\subsection{Our Approach vs. Traditional Sequential Routing}

As mentioned earlier, to simplify the problem and verify the proposed approach, this work is limited to single-layer circuits with one  optimization goal which is minimizing the total length of paths. The most popular existing solutions to this simplified problem are the sequential  maze 
routing in A* and Lee's algorithms~\cite{4082128,5219222}. 
These methods are known to suffer from the net ordering issues~\cite{zhang2016study}.
Among the 30 circuits used in our experiments, 20\% of them cannot be routed by traditional sequential routing based on A* or Lee's algorithm. 
In contrast, our approach successfully routes all of them.

\begin{figure*}[htbp]
\centering
\begin{subfigure}{0.34\textwidth}
  \centering
  \includegraphics[width=.95\textwidth]{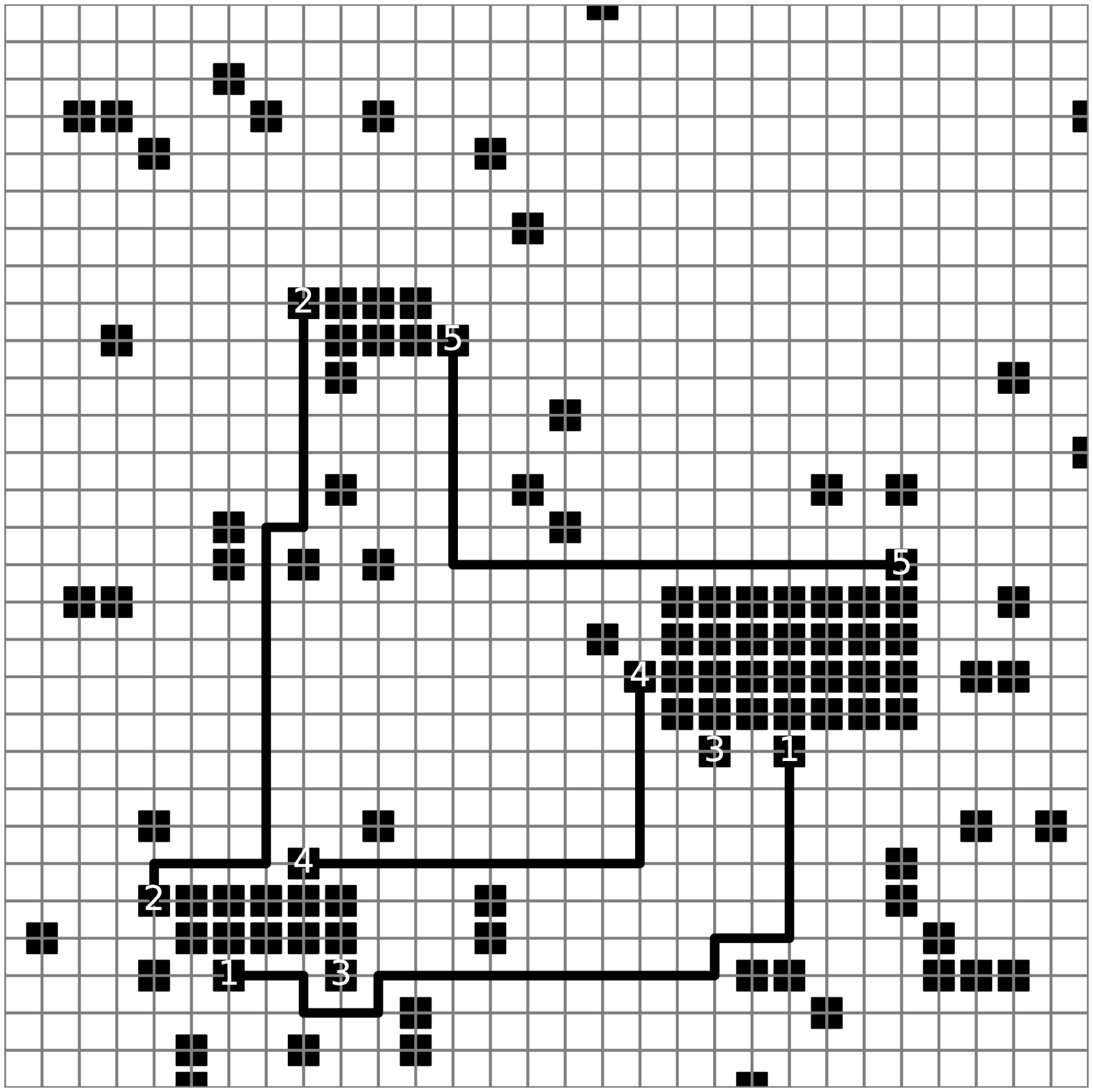}
  \caption{In Lee's algorithm. Failed. }
  \label{fig:route1}
\end{subfigure}%
\begin{subfigure}{0.34\textwidth}
  \centering
  \includegraphics[width=.95\textwidth]{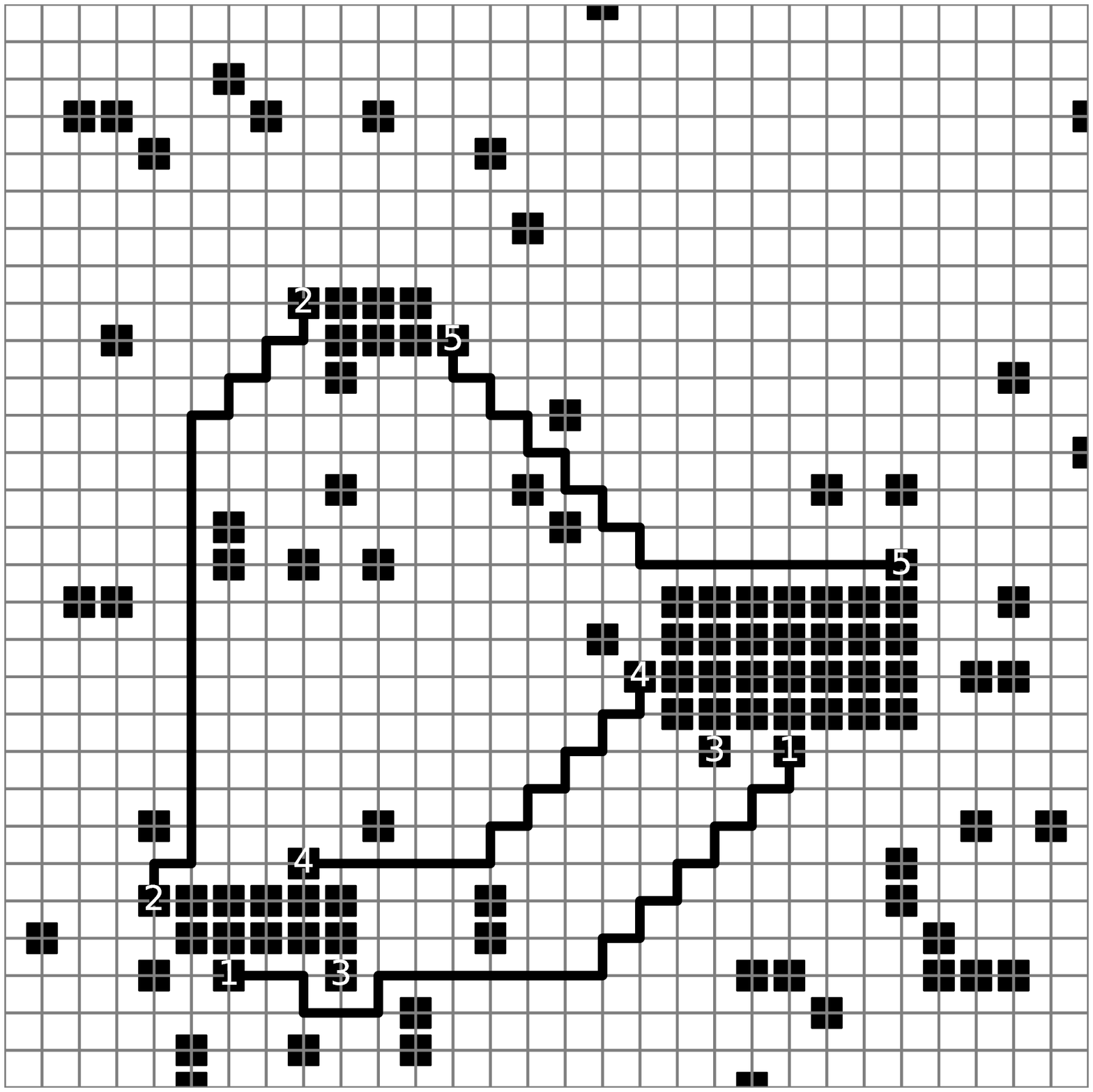}
  \caption{In A* algorithm. Failed. }
  \label{fig:route2}
\end{subfigure}%
\begin{subfigure}{0.34\textwidth}
  \centering
  \includegraphics[width=.95\textwidth]{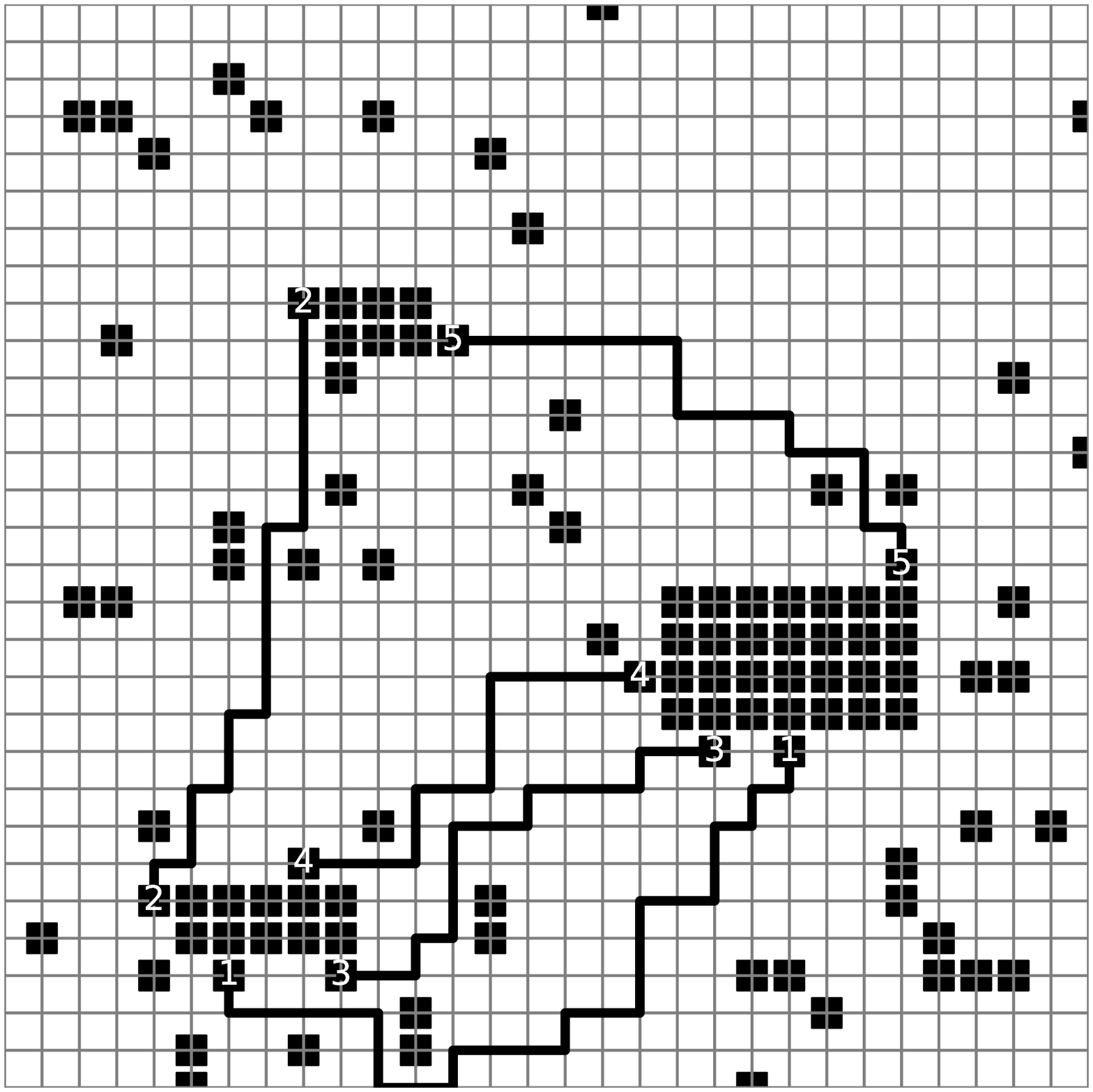}
  \caption{In our approach. Succeed. }
  \label{fig:route3}
\end{subfigure}
\caption{Routing results in our approach and traditional sequential routing. Obstacles are in solid black squares, nets are in numbered black squares, routed paths are in solid black lines. }
\label{fig:Routedboard}
\end{figure*}

Figure~\ref{fig:Routedboard} shows an example of routing in our approach and traditional sequential routing.
In Figures~\ref{fig:route1} and \ref{fig:route2}, 
traditional routing approaches connects nets 1, 2, 4 and 5 with shortest paths but makes it impossible to connect net $3$ because all possible solutions are blocked by the wire connecting net $1$.
Although still a sequential routing approach, our approach can leave space for nets to be routed later, instead of blocking them. 
As shown in Figure~\ref{fig:route3}, our approach lets the path of net 1 take a longer detour to leave space for the path of net 3 and results in a successful routing for all 5 nets. 


\subsection{Our Approach vs. Vanilla MCTS}
As discussed in Sections~\ref{sec:actionEva} and \ref{sec:rollout_policy}, vanilla MCTS is impractical in circuit routing. 
Here we will verify our improvements. 

To verify the effectiveness of DNN-DFS rollout policy, we compare the success rate of routing between vanilla MCTS and our approach in Table~\ref{Table:rollout_policy}. 
In the random rollout policy of vanilla MCTS, none of 30 circuits are successfully routed with either Avg-UCT or Max-UCT. All the routing processes go to dead ends. 
As mentioned in Section~\ref{sec:rollout_policy}, due to the large search space, random rollouts have little chance to reach successful routing terminal states despite their existence. 
In the DNN-DFS rollout policy, all nets can be successfully connected  within 1000 iterations. 


\setlength\tabcolsep{1.5pt} 
\begin{table}[!htbp]
\centering
\caption{The rate of successful routing ($\%$) by different numbers of  iterations
}
\begin{tabular}{@{}cccccccc@{}}
\toprule
\multirow{2}{*}{Rollout} & \multirow{2}{*}{UCT}
& \multicolumn{6}{c}{Number of iterations}    \\ \cmidrule(l){3-8} 
\multicolumn{2}{c}{}                                    & 100   & 200   & 500   & 1000   & 2000   & 5000   \\ \midrule
\multicolumn{1}{c}{\multirow{2}{*}{Random}}  & Avg. & 0.00  & 0.00  & 0.00  & 0.00   & 0.00   & 0.00   \\ 
\multicolumn{1}{c}{}                                 & Max. & 0.00  & 0.00  & 0.00  & 0.00   & 0.00   & 0.00   \\ \midrule
\multicolumn{1}{c}{\multirow{2}{*}{DNN-DFS}} & Avg. & 76.67 & 86.67 & 93.33 & 100.00 & 100.00 & 100.00 \\ 
\multicolumn{1}{c}{}                                 & Max. & 80.00 & 83.33 & 96.67 & 100.00 & 100.00 & 100.00  \\ \bottomrule
\end{tabular}
\label{Table:rollout_policy}
\end{table}
\setlength\tabcolsep{6pt} 

To study the effectiveness of the Max-UCT over Avg-UCT, we compare the lengths of total wires generated by them. 
A metric, \emph{wire redundancy ratio}, is introduced to 
show how redundant wires generated by an algorithm are with respect to the theoretically shortest paths. 
It is defined as $\frac{L_r-L_s}{L_s}\times 100\%$, where $L_r$ and $L_s$ are the lengths of routed wires and theoretically shortest wires, respectively. 
The results are reported in Table~\ref{Table:PWLD}. 
Because Avg-UCT based vanilla MCTS needs substantially more iterations to route a circuit than the Max-UCT based one, 
wire redundancy ratios are reported only for circuits successfully routed by both.

\begin{table}[!htbp]
\centering
\caption{Mean wire redundancy ratio ($\%$) of all successful-routing circuits by different numbers of iterations
}
\begin{tabular}{@{}ccccccc@{}}
\toprule
\multirow{2}{*}{\begin{tabular}[c]{@{}c@{}}Action evalua-\\ tion policy\end{tabular}} & \multicolumn{6}{c}{Number of iterations} \\ \cmidrule(l){2-7} 
                                                    & 100 & 200 & 500 & 1000 & 2000 & 5000 \\ \midrule
Avg-UCT  & 96.3       & 86.3       & 76.3       & 82.7        & 82.5  & 76.7     \\ 
Max-UCT & 12.5       & 13.5       & 9.7       & 8.2        & 6.9    & 4.5   \\ \bottomrule
\end{tabular}
\label{Table:PWLD}
\end{table}

The MCTS algorithm with Max-UCT generates much less redundant 
wires (wire redundancy ratios ranging from 12.5\% to 4.5\%) than that with Avg-UCT (always above 75\%). 
On top of that, MCTS with Max-UCT monotonically generates 
less redundant wires as the number of iterations increases, 
whereas MCTS with Avg-UCT does not show such a monotonic trend. 
As defined in Eqn.~(\ref{eqn:reward}), the rewards of a successful routing and a failed routing are different. Therefore, in the Avg-UCT mechanism, if a node in the search tree has a failed-routing experience, it needs more successful-routing rollouts to compensate for getting a good average reward value. As a result, it is quite possible that searching with more iterations leads a lower wire length redundancy value.

\section{Discussions and Limitations}\label{sec:limitations}
The limited but promising results show the potential of DNNs and tree search methods in circuit routing. 
First, by training with well-routed circuits, the CNN-based action prediction can learn a relatively satisfying routing strategy and provide a good guide for unseen circuits. 
Second, given enough iterations, the tree search with a trained policy can guarantee the success of routing and achieving optimization goals. 
Although our results are based on simple settings, the approach can be easily extended to the complex cases with more training data and more iterations of the tree search.

The limitation of this work is that the routing agent cannot learn and memorize the new strategy while routing new, unseen circuits. 
To solve this problem, we will combine reinforcement learning with tree search to facilitate the agent in learning new strategy while routing new, unseen circuits. 
In this way, as the agent processes more and more circuits, the policy will be improved. 
Furthermore, there is room to improve the search efficiency of MCTS by preserving the paths from each search tree. 
In current version of MCTS, the knowledge learned from calling MCTS at a state can not benefit the search at another state. 

\section{Conclusions}\label{sec:conclusion}
In this paper, we propose an end-to-end solution to circuit routing by combining deep neural networks (DNNs) and Monte Carlo tree search (MCTS).
Compared with traditional circuit routing approaches that are laborious to develop, inflexible to adapt, and difficult to maintain, 
our approach shifts the heavy lifting to 
the computer which searches over a massive space of routing solutions without human input. 
Given enough iterations, a solution is guaranteed  to be found if it exists. 
This approach can be easily implemented and adapted to any design specifications without changing the implementation code. 
Because vanilla MCTS is too slow for the large search space of circuit routing, two improvements are made. 
First, to deal with the large search space, we introduce DNN-guided DFS as the rollout policy. 
Second, to optimize on design goals, the maximum reward-based UCT is implemented in selection stage. 
Experimental results show that our approach can solve problems that baseline approaches cannot. 
They also show that the two improvements can greatly speed up the search and achieve more optimized results. 
Although our experiments are on single-layer circuits with minimizing total wire length as the only optimization goal, it can be extended to more complex designs easily, including but limited to more layers, considering via cost, subjecting to  line width and spacing specifications, etc.

\bibliographystyle{named}
\bibliography{ijcai20}

\end{document}